\documentclass[10pt,twocolumn,letterpaper]{article}

\usepackage{cvpr}
\usepackage{times}
\usepackage{epsfig}
\usepackage{graphicx}
\usepackage{amsmath}
\usepackage{amssymb}
\usepackage{multirow}
\usepackage{subfigure}

\usepackage[breaklinks=true,bookmarks=false]{hyperref}

\cvprfinalcopy 

\def\cvprPaperID{2104} 

\begin{document}

\title{SPM-Tracker: Series-Parallel Matching for Real-Time Visual Object Tracking}

\author{
\vspace{4pt}Guangting Wang$^{*1}$ \qquad Chong Luo$^{2}$ \qquad Zhiwei Xiong$^{1}$ \qquad Wenjun Zeng$^{2}$ \\
University of Science and Technology of China$^{1}$ \qquad Microsoft Research Asia$^{2}$ \\
{\tt\small wgting96@gmail.com\quad cluo@microsoft.com\quad zwxiong@ustc.edu.cn\quad wezeng@microsoft.com} 
}

\maketitle

\begin{abstract}

The greatest challenge facing visual object tracking is the simultaneous requirements on robustness and discrimination power. In this paper, we propose a SiamFC-based tracker, named SPM-Tracker, to tackle this challenge. The basic idea is to address the two requirements in two separate matching stages. Robustness is strengthened in the coarse matching (CM) stage through generalized training while discrimination power is enhanced in the fine matching (FM) stage through a distance learning network. The two stages are connected in series as the input proposals of the FM stage are generated by the CM stage. They are also connected in parallel as the matching scores and box location refinements are fused to generate the final results. This innovative series-parallel structure takes advantage of both stages and results in superior performance. The proposed SPM-Tracker, running at 120fps on GPU, achieves an AUC of 0.687 on OTB-100 and an EAO of 0.434 on VOT-16, exceeding other real-time trackers by a notable margin.
    
\end{abstract}
\footnote{$^{*}$This work is done when Guangting Wang is an intern in MSRA.}

\section{Introduction}
    
Visual object tracking is one of the fundamental research problems in computer vision and video analytics. Given the bounding box of a target object in the first frame of a video, a tracker is expected to locate the target object in all subsequent frames. The greatest challenge of visual tracking can be attributed to the simultaneous requirements on robustness and discrimination power. The robustness requirement demands a tracker not to lose tracking when the appearance of the target changes due to illumination, motion, view angle, or object deformation. Meanwhile, a tracker is required to have the power to discriminate the target object from cluttered background or similar surrounding objects. These two requirements are sometimes contradictory and hard to be fulfilled at the same time. 
        
Intuitively, both requirements need to be handled through online training. A tracker keeps collecting positive and negative samples along the tracking process. For generative trackers, positive samples help to model the appearance variation of the target. For discriminative trackers, more positive and negative samples help to find a more precise decision boundary that separates the target from the background. For quite a long time, online training has been an indispensable part in tracker design. Recently, with the advancement of deep learning and convolutional neural networks, deep features have been widely adopted in object trackers \cite{ nam2016learning, danelljan2016beyond, song2017crest, guo2017learning, danelljan2017eco, lu2018deep}. However, online training with deep features is extremely time consuming. Without much surprise, the deep version of many high-performance trackers \cite{danelljan2016beyond, danelljan2017eco, bhat2018unveiling, song2017crest, nam2016learning, wang2018multi, yao2018joint} cannot run in real-time any more, even on modern GPUs.

\begin{figure}[t]
    \begin{center}
        \includegraphics[width=0.90\columnwidth]{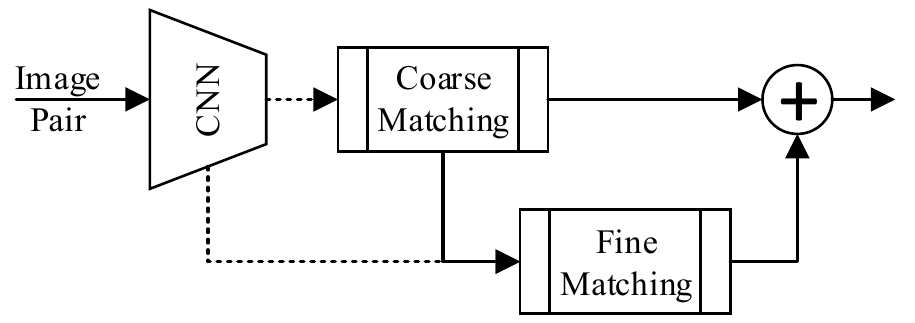}
    \end{center}
    \caption{The series-parallel structure which connects coarse matching and fine matching stages in the proposed SPM-Tracker.}
    \label{fig:sp_diagram}
    \vspace{-6mm}
\end{figure}
    
While the excessive volume of deep features brings speed issues to online training, their strong representational power also opens up a possibility to completely give up online training. The idea is, under a given distance measure, to learn an embedding space, through offline training, that can maximize the interclass inertia between different objects and minimize the intraclass inertia for the same object \cite{zhu2018distractor}. Note that maximizing the interclass inertia corresponds to the discrimination power and minimizing the intraclass inertia corresponds to the robustness. The pioneering work along this research line is SiamFC \cite{bertinetto2016fully}. In addition to the offline training, SiamFC uses cross-correlation operation to efficiently measure the distance between the target patch and all surrounding patches. As a result, SiamFC can operate at 86fps on GPU. 

By design, the SiamFC framework faces challenges in balancing the robustness and the discrimination power of the embedding and in handling the scale and aspect ratio changes of the target object. Recently, SiamRPN \cite{li2018high} was proposed to address the second challenge. It consists of a Siamese subnetwork for feature extraction and a region proposal subnetwork for similarity matching and box regression. In a follow-up work called DaSiamRPN \cite{zhu2018distractor}, distractor-aware training is adopted to promote the generalization and discriminative ability of the embedding. In these two pieces of work, visual object tracking is formulated as a local one-shot detection task.

In this paper, we design a two-stage SiamFC-based network for visual object tracking, aiming to address both challenges mentioned above. The two stages are the coarse matching (CM) stage which focuses on enhancing the robustness and the fine matching (FM) stage which focuses on improving the discrimination power. By decomposing these two equally important but somewhat contradictory requirements, our proposed network is expected to achieve better performance. Moreover, both CM and FM stages perform similarity matching and bounding box regression. Thanks to the two-stage box refinement, our tracker achieves high localization precision without multi-scale test.

The key innovation in this work is the series-parallel structure that is used to connect the two stages. The schematic diagram is shown in Fig.\ref{fig:sp_diagram}. Similar to the series structure which is widely adopted in two-stage object detection, the input of the second FM stage relies on the output of the first CM stage. In this sense, the CM stage is a proposal stage. Similar to the parallel structure, the final matching score as well as the box location are the fused results from both stages. This series-parallel structure brings a number of advantages which will be detailed in Section \ref{section:approach}. In addition, we propose generalized training (where objects from the same category are all treated as the same object) to boost the robustness of the CM stage. The discrimination power of the FM stage is promoted by replacing the cross-correlation layer with a distance learning subnetwork. With these three innovations, the resulting tracker achieves superior performance on major benchmark datasets. It achieves an AUC of 0.687 on OTB-100 and EAOs of 0.434 and 0.338 on VOT-16 and VOT-17, respectively. More importantly, the inference speed is 120fps on a NVIDIA P100 GPU. 

The rest of the paper is organized as follows. We discuss related work in Section \ref{section:related}. The proposed series-parallel framework is presented in Section \ref{section:approach}. After describing the implementation details of SPM-Tracker in Section \ref{section:implementation}, we provide extensive experimental results in Section \ref{section:experiments}. Finally, we conclude the paper with some discussions in Section \ref{section:conclusion}.

\begin{figure*}[t]
    \begin{center}
    \includegraphics[width=0.9\linewidth]{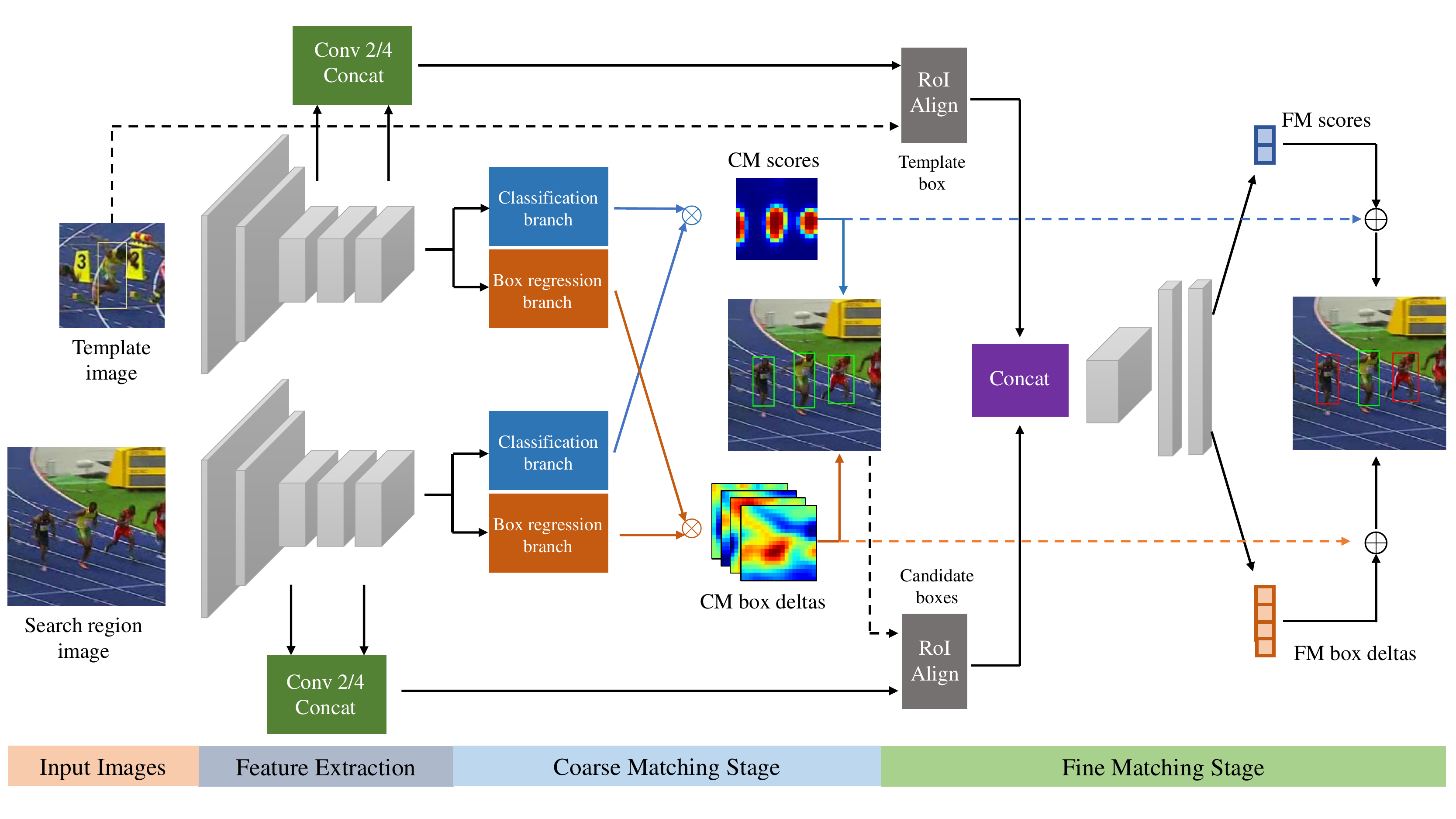}
    \end{center}
    \vspace{-6mm}
       \caption{Details of the proposed series-parallel matching framework. We employ Siamese AlexNet \cite{krizhevsky2012imagenet} for feature extraction. The CM stage adopts the network structure of SiamRPN \cite{li2018high}. RoI Align \cite{he2017mask} is used to generate fixed-length regional features for each proposal. The FM stage implements a relation network \cite{yang2018learning} for distance learning. Finally, results from both stages are fused for decision making. }
    \label{fig:framework}
    \vspace{-4mm}
\end{figure*}

\section{Related Work} \label{section:related}

Object trackers have conventionally been classified into generative trackers and discriminative trackers \cite{kalal2011TLD}, and most modern trackers belong to the latter. A common approach of discriminative trackers is to build a binary classifier that represents the decision boundary between the object and its background \cite{kalal2011TLD}. It is generally believed that adaptive discriminative trackers, which continuously update the classifier during tracking, are more powerful than their static counterparts. 


Correlation Filter (CF) based trackers are among the most successful and representative adaptive discriminative trackers. Bolme \textit{et al.} \cite{bolme2010visual} first proposed the MOSSE filter which is capable of producing stable CFs from a single frame and then continuously being improved during tracking. The MOSSE filter has aroused a great deal of interest and there are a bunch of follow-up work. For example, kernel tricks \cite{henriques2012exploiting, henriques2015high, danelljan2014adaptive} were introduced to extend CF. DSST \cite{danelljan2014adaptive} and SAMF \cite{li2014scale} enabled scale estimation in CF. SRDCF \cite{danelljan2015learning} was proposed to alleviate the periodic effect of convolution boundaries.

More recently, with the advancement of deep learning, the rich representative power of deep features is widely acknowledged. There is a trend to utilize deep features in CF-based trackers \cite{ma2015hierarchical, danelljan2016beyond, danelljan2017eco, bhat2018unveiling}. However, this creates a dilemma: online training is an indispensable part of CF-based trackers, but online training with deep features is extremely slow. As a result, the high-speed signature of CF-based trackers is fading.

In many real world applications, being real-time is mandatory for a tracker. Facing the above mentioned dilemma, many researchers resorted to another choice: static discriminative trackers. With the highly expressive deep features, it becomes possible to build high-performance static trackers. This idea was successfully realized by SiamFC \cite{bertinetto2016fully}. SiamFC employs Siamese convolutional neural networks (CNNs) to extract features, and then uses a simple cross-correlation layer to perform dense and efficient sliding-window evaluation in the search region. Every patch of the same size as the target gets a similarity score, and the one with the highest score is identified as the new target location. There are also a great number of follow-up work \cite{guo2017learning, yang2018mem, wang2018learning}, among which SA-Siam \cite{he2018twofold, he2018towards} and SiamRPN \cite{li2018high, zhu2018distractor} are most related to ours. 

The main challenge in SiamFC-based methods is to find an embedding space, through offline training, that is both robust and discriminative. Zhu et al. \cite{zhu2018distractor} propose distractor-aware training to emphasize these two aspects. They use diverse categories of positive still image pairs to promote the robustness, and use semantic negative pairs to improve the discriminative ability. However, it is  difficult to attend to both requirements in a single network. SA-Siam \cite{he2018twofold} and Siam-BM \cite{he2018towards} adopt a two-branch network to encode images into two embedding spaces, one for semantic similarity (more robust) and the other for appearance similarity (more discriminative). This typical parallel structure does not take advantage of the innate proposal capability of the semantic branch.

Another challenge in SiamFC-based methods is how to handle scale and shape changes. Almost all SiamFC-based trackers adopt an awkward multi-scale test for scale adjustment, but the aspect ratio of bounding boxes remains unchanged throughout the tracking process. SiamRPN \cite{li2018high} addresses this issue with an elegant region proposal network (RPN). The capability to do box refinement also allows it to discard multi-scale test. In this work, we follow SiamRPN to use RPN for bounding box size adjustment. The two-stage refinement allows our SPM-Tracker to achieve an even more precise box location.

SiamRPN and DaSiamRPN \cite{zhu2018distractor} pose the tracking problem as a local single-stage object detection problem. Some recent empirical studies \cite{huang2017speed} on object detection show that two-stage design is often more powerful than one-stage design. This may be related to hard example mining \cite{lin2017focal} and regional feature alignment \cite{he2017mask}. In the tracking community, Zhang \textit{et al.} \cite{zhang2018LRVNT} adopt a two-stage design for long-term tracking. However, the series structure they adopted demands for a very powerful second stage. They use MDNet \cite{nam2016learning} for the second stage, which greatly slows down the inference speed to 2fps. 


\section{Our Approach} \label{section:approach}
    
\subsection{Series-Parallel Matching Framework} \label{section: approach_overview}

We propose a framework for robust and discriminative visual object tracking. The proposed SPM framework is depicted in Fig.\ref{fig:framework}. We employ a Siamese network to extract features from the target patch and the local search region. This is followed by two matching stages, namely coarse matching stage and fine matching stage, organized in a series-parallel structure. 

Both the CM and FM stages produce similarity scores of proposals and box location deltas. We let the CM stage to focus on the robustness, i.e. to minimize the intraclass inertia for the same object. It is expected to propose the target object even when it is experiencing huge appearance changes. A number of proposals which get the top matching scores in the CM stage are then passed to the FM stage and fixed-size regional features are extracted through RoI Align \cite{he2017mask}. The FM stage is designed to focus on discrimination, i.e. to maximize the interclass inertia between different objects. It is expected to discriminate the true target from background or surrounding similar objects. Eventually, the matching scores and box locations from both matching stages are fused to make the final decision. 

The proposed SPM framework brings a number of advantages as outlined below. 
\begin{itemize}
\vspace{-2mm}\item The robustness and the discrimination requirements are decomposed and emphasized in separate stages. It is easier to train two networks to achieve their respective goals than to train a single network that simultaneously achieves the goals for both requirements.. 
\vspace{-2mm}\item The input proposals of the FM stage are all high-score candidates from the CM stage. FM stage training benefits from a balanced positive-negative ratio and hard negative mining to enhance the discrimination power. 
\vspace{-2mm}\item Box regression in the CM stage allows the FM stage to evaluate aligned patches with different scale or even different aspect ratio from the target object. Fusion of two-stage box regressions leads to a higher precision. 
\vspace{-2mm}\item Since only a few proposals are passed to the FM stage, it is not necessary to use cross-correlation operation to compute distance. We could adopt a trainable distance measure for the FM stage.
\end{itemize}

In the following two subsections, we will discuss the CM and FM stages in more details. 

    \begin{figure}[t]
        \begin{center}
            \includegraphics[width=0.90\linewidth]{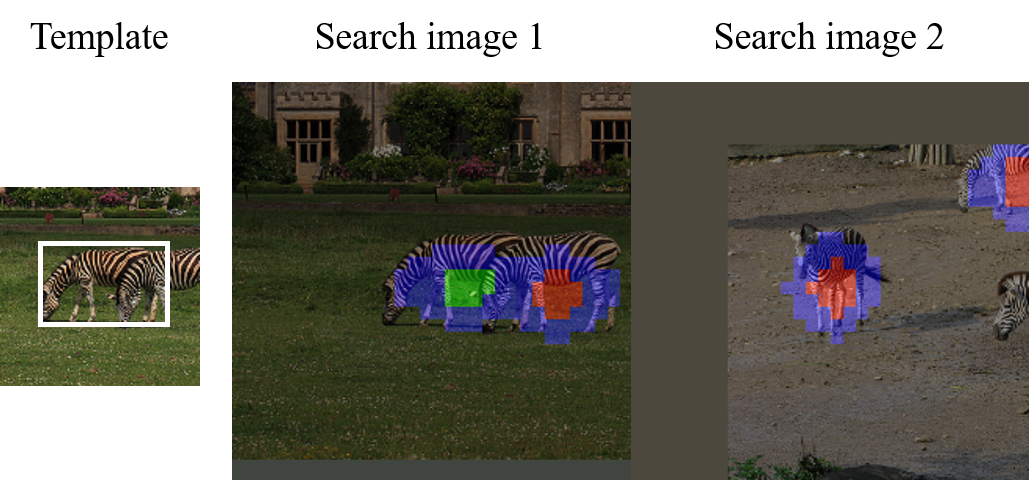}
        \end{center}
        \vspace{-3mm}
        \caption{Illustration of the generalized training (GT) strategy for the CM stage. Given a template as shown on the left, the green blocks in search image 1 indicate the positive samples used in conventional training. The red blocks are the locations of other objects of the same category. GT takes both green and red blocks as positive samples. (The blue blocks indicate the ignored region.) Best viewed in color.}
        \label{fig:training_scheme}
        \vspace{-2mm}
    \end{figure}

\begin{figure}[]
    \begin{center}
        \includegraphics[width=0.90\linewidth]{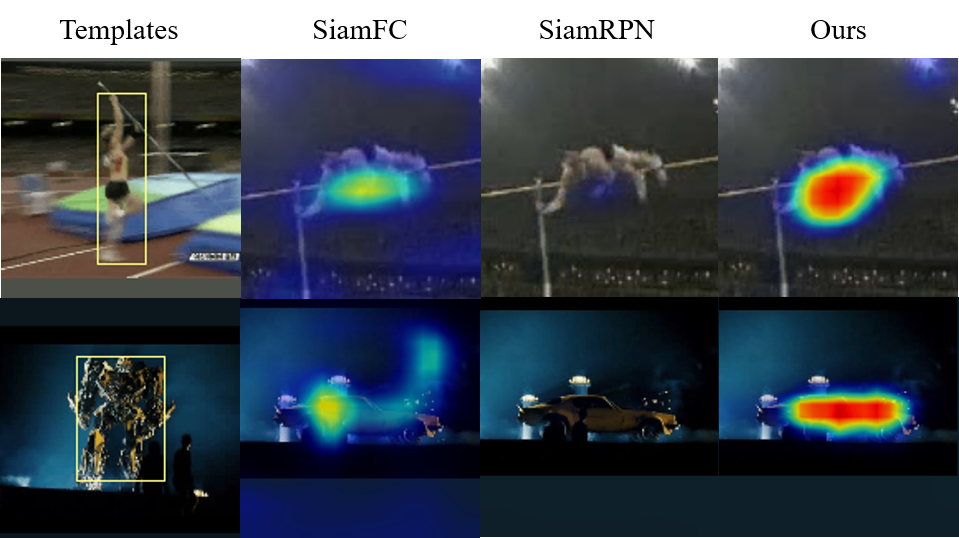}
    \end{center}
    \vspace{-3mm}
    \caption{Visualization of the cross-correlation response maps generated by SiamFC \cite{bertinetto2016fully}, SiamRPN \cite{li2018high}, and the CM stage of our tracker. Our tracker can robustly highlight the target object even when it has severe deformation. Best viewed in color.}
    \label{fig:robustness}
    \vspace{-4mm}
\end{figure}

\subsection{Coarse Matching Stage} \label{section:approach_cm}

The coarse matching stage looks in the search region for candidate patches which are similar to the target patch. It is expected to be very robust such that the target object will not be missed even when it is experiencing drastic appearance changes due to intrinsic or extrinsic factors. We adopt the region proposal subnetwork as introduced in SiamRPN \cite{li2018high} for this stage. Given the features extracted by a Siamese network, pair-wise correlation feature maps are computed for the classification branch and the regression branch. The classification branch produces the similarity scores for the candidate boxes while the regression branch generates the box deltas. Similar to SiamRPN, we can discard multi-scale test since the proposal network handles scale and shape changes in a graceful manner. 

For the CM stage, we propose generalized training (GT) to improve the robustness. Conventionally, image pairs of the same object drawn from two frames of a video are used as positive samples. In DaSiamRPN \cite{zhu2018distractor}, still images from detection datasets are used to generate positive image pairs through data augmentation. In this work, we additionally treat some image pairs containing different objects as positive samples when the two objects belong to the same category. Fig. \ref{fig:training_scheme} illustrates the classification labels used in our CM stage and in other SiamFC-based trackers. This training strategy leads to exceedingly generalized embeddings which capture high-level semantic information and therefore are insensitive to object appearance changes.

Fig.\ref{fig:robustness} shows the response map of the CM stage and compares it with that of SiamFC and SiamRPN (with distractor-aware training). It is observed that our tracker is able to generate strong responses even when the target object has significant deformation. By contrast, SiamRPN \cite{li2018high, zhu2018distractor} barely produce any response and SiamFC does not have a precise localization. 

\begin{figure}[]
    \begin{center}
        \includegraphics[width=0.90\linewidth]{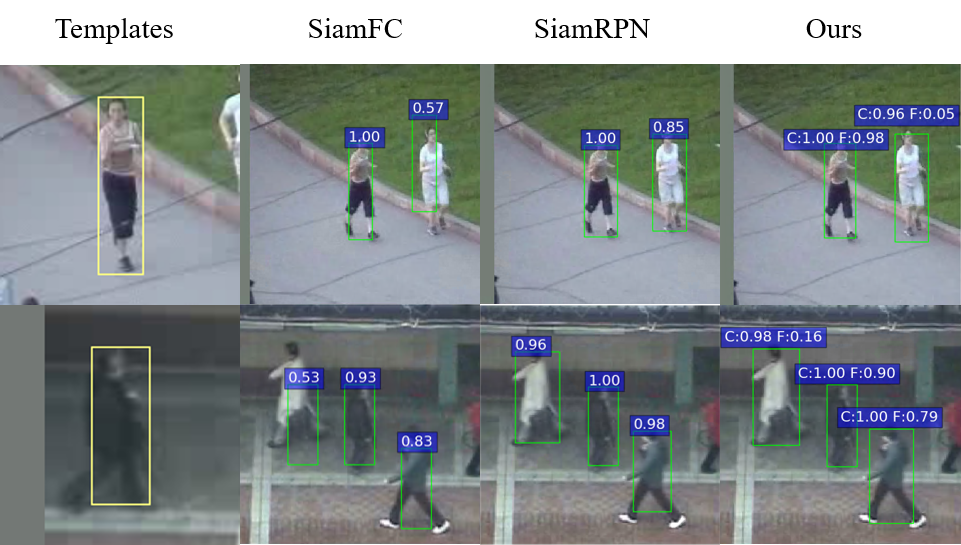}
    \end{center}
    \vspace{-3mm}
    \caption{Visualization of the top-K matched boxes and their similarity scores output by SiamFC \cite{bertinetto2016fully}, SiamRPN \cite{li2018high}, and our SPM-Tracker. Our tracker generates two scores, corresponding to the CM stage (\textbf{C}) and the FM stage (\textbf{F}). Objects of the same category get high C-scores but only the true target gets high F-scores. It shows that SPM-Tracker achieves the design goal.}
    \label{fig:discrimination}
    \vspace{-5mm}
\end{figure}
    
\subsection{Fine Matching Stage} \label{section:approach_fm}

The fine matching stage is expected to capture fine-grained appearance information so that the true target can be distinguished from background or similar surrounding objects. The FM stage only evaluates the top $K$ highest-score patches from the CM stage.

As illustrated in Fig. \ref{fig:framework}, the FM stage shares features with the CM stage. For each proposal, the regional features are directly cropped from the shared feature maps. Considering the fact that shallow features contain detailed appearance information and also result in high localization precision, we take both deep and shallow features and fuse them by concatenation. Then, RoI Align operation \cite{he2017mask} creates fixed-size feature maps for each proposal. 

Since there are only a limited number of patches to be evaluated in this stage, we can afford to use a more powerful distance learning network, instead of the cross-correlation layer, to measure the similarity. Additionally, such a network could be trained to generate a complementary score to the CM similarity scores. We adopt a light-weight relation network as proposed in \cite{yang2018learning} for the FM stage. The input of the relation network is the concatenated feature from the image pairs. A $1 \times 1$ convolution layer is followed by two fully connected layers which generate feature embedding for classification and box regression. 

Finally, the similarity scores and the box deltas from two stages are fused by weighted sum. The candidate box with the highest similarity score is identified as the target object. Fig. \ref{fig:discrimination} shows the top-K candidates and their similarity scores output by different trackers. Our tracker is associated with two scores corresponding to the CM and FM stages. The high C-scores for all the foreground objects suggest the robustness of SPM-Tracker and the low F-scores for non-target objects demonstrate the discrimination power.

\section{Implementation} \label{section:implementation}
    
\subsection{Network Structure and Parameters} \label{section:implementation_architecture} 
    
The CNN backbone used for feature extraction is the standard AlexNet \cite{krizhevsky2012imagenet}. It is pre-trained on the ImageNet dataset. Unlike other SiameFC-based trackers, we keep the padding operations in the backbone network. This is because the RoI Align operation needs pixel alignment between feature maps and source images. The CM stage still uses the central features without padding. In our implementation, the target image has a size of $127 \times 127 \times 3$. The size of its last-conv-layer feature map with padding is $16 \times 16 \times 256$. Only the central $6 \times 6 \times 256$ features are used for the CM stage, which is consistent with the original SiamFC. The FM stage extracts regional features from \textit{conv2} (384 channels) and \textit{conv4} (256 channels) layers and concatenates them. We use RoI Align operation to pool regional features of size $6 \times 6 \times 640$ for each proposal, where $6 \times 6$ is the spatial size and 640 is the number of channels. The two fully-connected layers in the FM stage are lightweight, with only 256 neurons per layer. 


\subsection{Training} \label{section:implementation_training} 
    
The entire network can be trained end-to-end. The overall loss function is composed of four parts: classification loss and box regression loss in both the CM stage and FM stage. For the CM stage, an anchor box is assigned a positive (or negative) label when its intersection-over-union (IoU) overlap with the ground-truth box is greater than 0.6 (or less than 0.3). Other patches whose IoU overlap falls in between are ignored. For the FM stage, positive (or negative) labels are assigned to candidate boxes whose IoU overlaps are greater (or less) than 0.5. Same as in the Faster R-CNN object detection framework \cite{ren2015faster}, box regression loss is added to positive samples in both stages. We adopt cross-entropy loss for classification and smooth L1 loss \cite{girshick15fastrcnn} for box regression. The overall loss function can be written as:
\begin{equation} \label{eq:all_loss}
    \begin{split}
      L &= \lambda_{1}L_{cm\_cls} + \lambda_{2}L_{cm\_b} + \lambda_{3}L_{fm\_cls} + \lambda_{4}L_{fm\_b},
    \end{split}
\end{equation}

\noindent where $L_{cls}$ denotes the classification loss and $L_{b}$ denotes the box regression loss. We set $\lambda_{2}=2$ and $\lambda_{1}=\lambda_{3}=\lambda_{4}=1$ since the box regression loss of CM module is much smaller than the others. 

The training image pairs are extracted from both videos and still images. The video datasets include VID \cite{russakovsky2015imagenet} and the training set of Youtube-BB \cite{real2017youtube}. Following DaSiamRPN \cite{zhu2018distractor}, we also make use of still image datasets, including COCO \cite{lin2014microsoft}, ImageNet DET \cite{russakovsky2015imagenet}, Cityperson \cite{zhang2017citypersons} and WiderFace \cite{yang2016wider}. The sampling ratio between videos  and still images is $4:1$. There are three types of image pairs, denoted by same-instance, same-category, and different-category. They are sampled at a ratio of $2:1:1$. 

    
The standard SGD optimizer is adopted for training. In each step, the CM stage produces hundreds of candidate boxes, among which 48 boxes are selected to train the FM stage. The positive-negative ratio is set to $1:1$. The learning rate is decreased from $10^{-2}$ to $10^{-4}$. We train the network for 50 epochs and 160,000 image pairs are sampled in each epoch.

\subsection{Inference} \label{section:implementation_inference}

During inference, we crop the template image patch from the first frame and feed it to the feature extraction network. The template features are cached so that we do not need to compute it in the subsequent frames. 
    
Given the tracking box in the last frame, a search image patch surrounding the box location is cropped and resized to $271 \times 271$. The CM stage takes the search image as input and then outputs a number of boxes. The candidate box that has the largest overlap with the tracking box in the previous frame will be reserved to keep the stability. Other boxes go through the standard proposal processing in RPN \cite{ren2015faster}. First, boxes with low scores are filtered. Then non-maximum suppression (NMS) is applied. The NMS threshold is 0.5. Finally, $K$ candidate boxes with top scores are selected and passed to the FM stage. In this step, we do not add shape penalties or cosine window penalties in order to aggressively propose boxes. The number of candidate boxes $K$ is set to 9, which is further analyzed in Section \ref{section:experiment_cm}. We use five anchors whose aspect ratios are $[0.33, 0.5, 1.0, 2.0, 3.0]$.
    
In the FM stage, similarity scores and refined box positions are predicted by the classification head and the box regression head. Let $u_c,u_f$ be the scores predicted by CM and FM stages, respectively. Let $\mathbf{B_c}, \mathbf{B_f}$ be the bounding box locations after the adjustment of the CM and FM stages, respectively. The final score and box coordinates are the weighted sum of the results from the two modules:
\begin{equation} \label{eq:final_score}
    \begin{split}
       u &= (1-W_{cls})u_c + W_{cls}u_f \\
       \mathbf{B} &= \frac{u_c}{W_{box}u_f + u_c}\mathbf{B_c} + \frac{W_{box}u_f}{W_{box}u_f + u_c}\mathbf{B_f},
    \end{split}
\end{equation}
\noindent where $W_{cls}, W_{box}$ are weights of the FM module for similarity score and box coordinates. We find that good tracking results are usually achieved when $W_{cls}$ takes a value around 0.5 and $W_{box}$ takes a value of 2 or 3. 
    
After applying cosine windows \cite{bertinetto2016fully}, the candidate box with the highest score is selected and its size is updated by linear interpolation with the result in the previous frame. Our tracker can run inference at 120fps with a single NVIDIA P100 GPU and an Intel Xeon E5-2690 CPU. 

\section{Experiments} \label{section:experiments}

The three main contributions in this work are: 1) using series-parallel structure to connect two matching stages; 2) adopting generalized training for the CM stage; and 3) adopting a relation network for distance measurement in the FM stage. In this section, we will first perform ablation analyses which support our contributions, and then carry out comparison studies with the state-of-the-art trackers on major benchmark datasets. 

\subsection{Analysis of the Series-Parallel Structure} \label{section:experiments_ablation}

    \begin{table}[]
        \begin{center}
        \begin{tabular}{|c|ccc|}
        \hline
                & \begin{tabular}[c]{@{}c@{}}CM \\ Only\end{tabular} & \begin{tabular}[c]{@{}c@{}}CM+FM \\ Series \end{tabular}    & \begin{tabular}[c]{@{}c@{}}CM+FM \\ Series-Parallel \end{tabular} \\ \hline
        \begin{tabular}[c]{@{}c@{}}OTB-100 (AUC)\end{tabular} & 0.643                                              & 0.632 & \textbf{0.670}                                                            \\ \hline
        \begin{tabular}[c]{@{}c@{}}VOT-17 (EAO)\end{tabular}  & 0.279                                              & 0.296 & \textbf{0.323}                                                          \\ \hline
        \begin{tabular}[c]{@{}c@{}}VOT-16 (EAO)\end{tabular}  & 0.359                                              & 0.343 & \textbf{0.391}                                                        \\ \hline
        \end{tabular}
        \end{center}
        \caption{Ablation analysis of different architectures. Results on three benchmark datasets are consistent, and they demonstrate the superiority of the series-parallel structure.}
        \label{table:architecture_ablation}
        \vspace{-5mm}
    \end{table}
    
We corroborate the effectiveness of the series-parallel structure by comparing it with two alternatives. The baseline scheme, denoted by ``CM only'' in Table \ref{table:architecture_ablation} is actually SiamRPN \cite{li2018high}. Our implementation achieves a slightly better performance than reported in their original paper (0.279 vs 0.244 on VOT-17 benchmark) because we have included additional still images in the training. When the FM stage is added in series, the performance (denoted by ``CM+FM Series'' in Table \ref{table:architecture_ablation}) is similar to the baseline (better on VOT-17 and worse on OTB-100 and VOT-16). 

The proposed ``CM+FM Series-Parallel'' method, which performs two-stage fusion, significantly outperforms the other two schemes, as Table \ref{table:architecture_ablation} shows. 
The reason why fusion plays an important role is that the two stages pay attention to different aspects of tracker capabilities: robustness at the CM stage and discrimination power at the FM stage. The matching score produced by one stage does not reflect the other capability. The idea of fusion has been practiced in many trackers \cite{wang2015transferring, gao2014transfer, chen2014description, he2018twofold} and has shown effectiveness. 




\subsection{Analysis of the CM Stage} \label{section:experiment_cm}

    \begin{table}[]
    \begin{center}
    \begin{tabular}{|l|ccc|}
    \hline
                      & \multicolumn{1}{c}{\begin{tabular}[c]{@{}c@{}}OTB-100\\ AUC\end{tabular}} & \multicolumn{1}{c}{\begin{tabular}[c]{@{}c@{}}VOT-17\\ EAO\end{tabular}} & \multicolumn{1}{c|}{\begin{tabular}[c]{@{}c@{}}VOT-16\\ EAO\end{tabular}} \\ \hline
    S-P model         & 0.670                                                                      & 0.323                                                                     & 0.391                                                                     \\ \hline
    S-P model + GT & \textbf{0.687}                                                                      & \textbf{0.338}                                                                     & \textbf{0.434}                                                                     \\ \hline
    \end{tabular}
    \end{center}
    \caption{Generalized training (GT) for the CM stage significantly improves the performance.}
    \label{table:ablation_robust}
    \vspace{-5mm}
    \end{table}
    
    \begin{figure}[]
    \centering
        \subfigure[]{%
        \begin{minipage}{0.230\textwidth}
        \centering
        \includegraphics[width=1.0\textwidth]{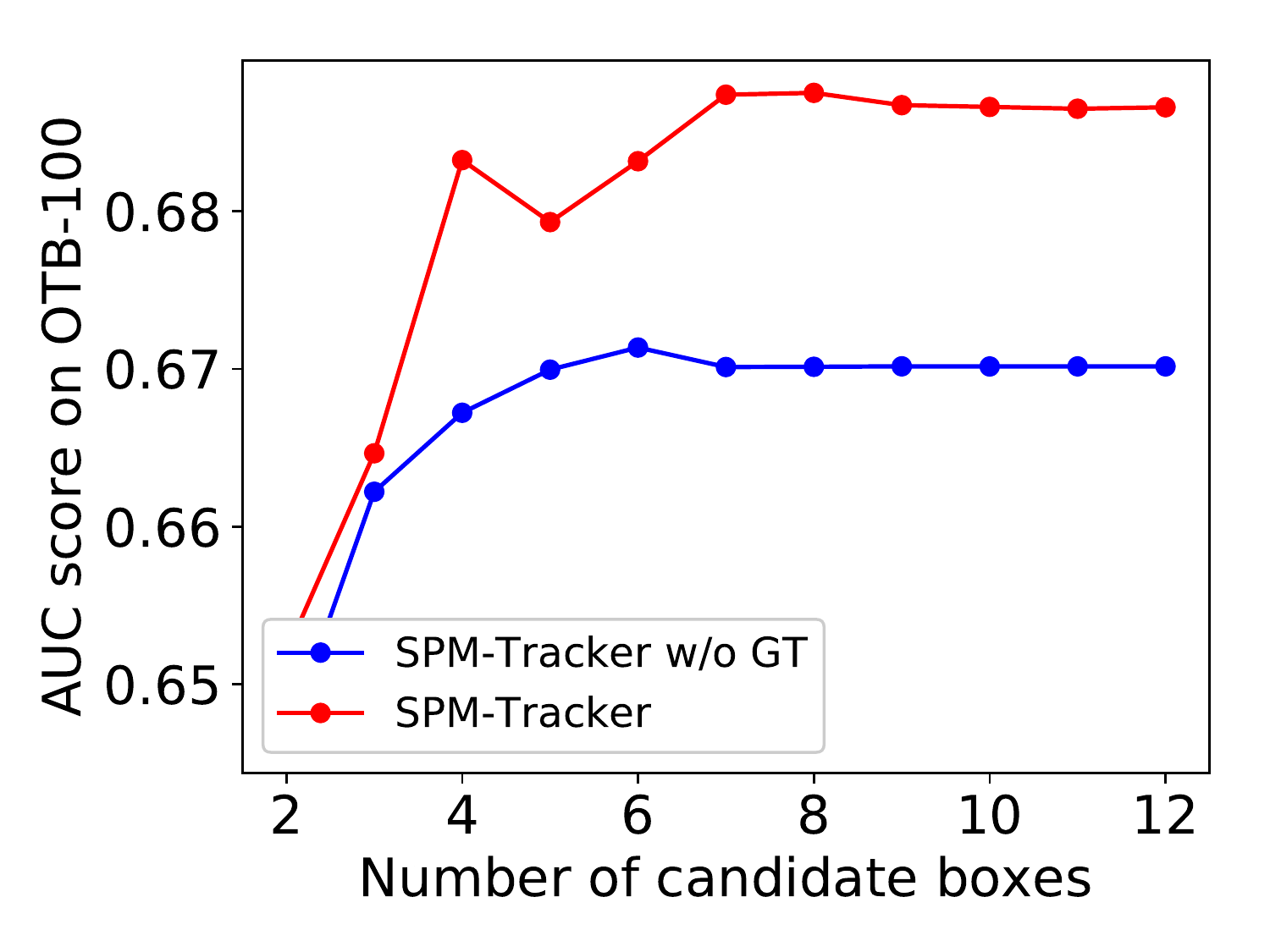}
         \end{minipage}
    }
        \subfigure[]{%
            \begin{minipage}{0.230\textwidth}
            \centering
            \includegraphics[width=1.0\textwidth]{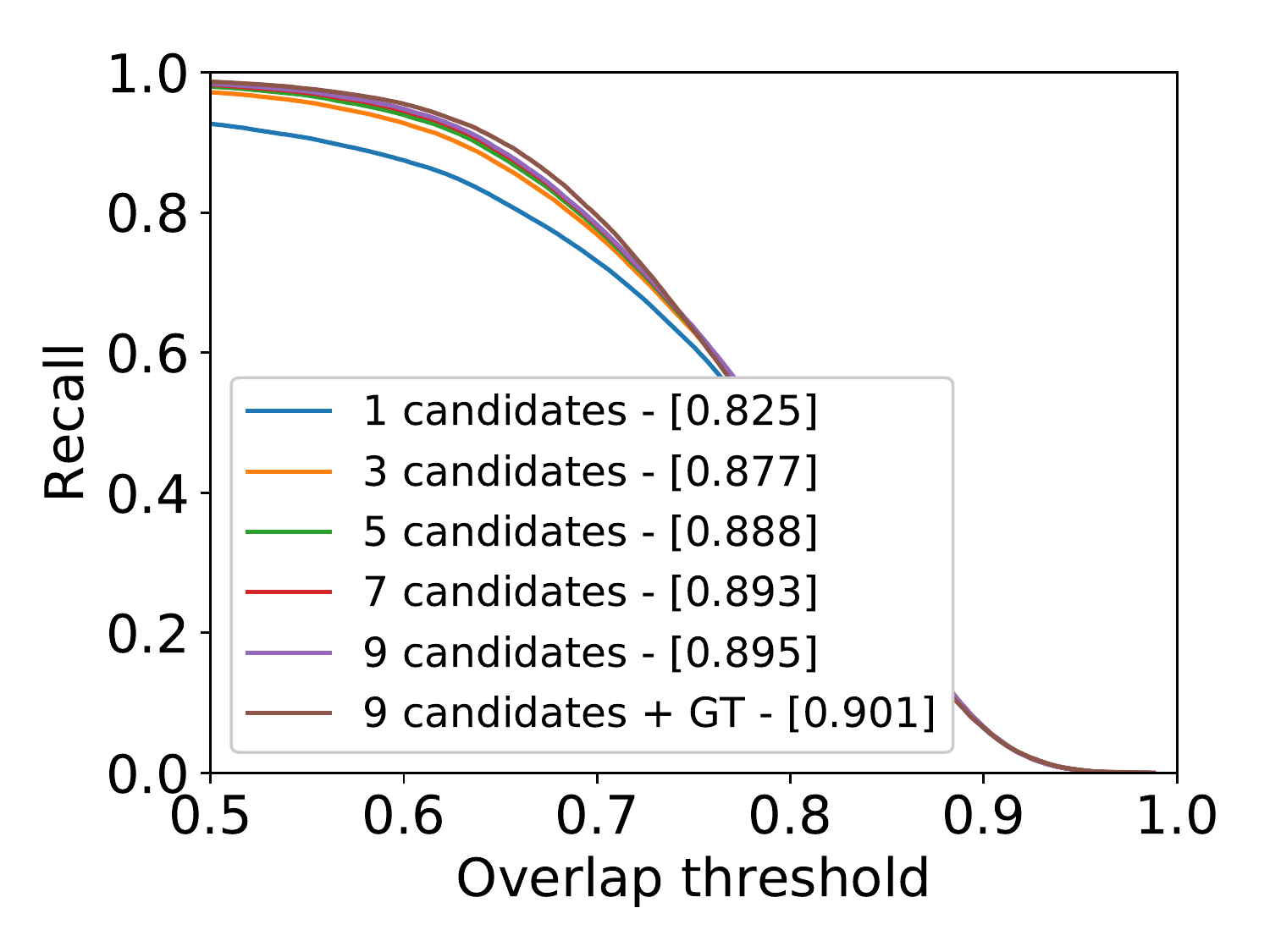}
             \end{minipage}
        }

    \caption{CM module analysis: (a) AUC score vs. number of candidate boxes; (b) recall rate vs. overlap thresholds (the values in brackets indicate the mean recalls over thresholds 0.5:0.05:0.7). All experiments are carried out on OTB-100 dataset.}
    \vspace{-5mm}
    \label{fig:proposal_module_analysis}
    \end{figure}

\noindent\textbf{Generalized Training Strategy:}
To make the CM module more robust to object appearance change, we propose to take image pairs in the same category as positive samples during training. This is referred to as the generalized training (GT) strategy. We compare the performance of SPM-Tracker when it is trained with or without the GT strategy for the CM stage. Improvements achieved on all three benchmark datasets, as shown in Table \ref{table:ablation_robust}, confirm the effectiveness of this strategy. Some of the visualization results have already been presented in Fig. \ref{fig:robustness} to show that the GT strategy helps to locate objects with large deformation. 

\noindent\textbf{Number of Candidate Boxes:} 
During inference, the CM stage passes $K$ top-scored candidate boxes to the FM stage. On the one hand, the larger the $K$ is, the higher the probability that the true target is included in the final evaluation. On the other hand, a larger $K$ means more false positives will be evaluated in the FM stage, which reduces the speed and might decrease the accuracy as well. In order to determine $K$, we investigate the relationship between the tracking performance and the number of candidate boxes. Fig. \ref{fig:proposal_module_analysis}(a) shows how the AUC on OTB-100 changes with $K$. We find that when $K$ is larger than 7, the performance tends to flatten. Therefore, we choose $K=9$ in experiments. 

\noindent\textbf{Recall:} 
Recall of candidate boxes can be used to measure the robustness. We use recall to further validate the GT strategy and $K$ selection in the CM stage. To ensure fairness, we crop the template from the first frame and the search region in the current frame is generated according to the ground-truth box in the previous frame. Fig. \ref{fig:proposal_module_analysis} (b) shows the recall vs. overlap threshold. The mean recall for the overlap thresholds $[0.5, 0.7]$ is also computed and listed in brackets. It is obvious that using a single candidate box results in significantly lower recall than using multiple candidates. As the number of candidate boxes increases, the recall also increases before it saturates at around $K=9$. At the saturation point, applying the GT strategy still can boost recall. This double confirms the power of the GT strategy. 

\begin{table}[]
\small
\setlength{\tabcolsep}{4pt}
\begin{center}
\begin{tabular}{|l|l|ccc|c|}
\hline
 & \multirow{2}{*}{Tracker} & \multicolumn{3}{c|}{AUC score (OPE)} & Speed  \\
 &                          & OTB-2013    & OTB-50    & OTB-100   &  (FPS) \\ \hline
 \multirow{8}{*}{\rotatebox{90}{CF-based Trackers}}
& LCT \cite{ma2015long}                          & 0.628       & 0.492     & 0.568     &27    \\
& Staple \cite{bertinetto2016staple}             & 0.593       & 0.516     & 0.582     &80    \\
& LMCF \cite{wang2017large}                      & 0.628       & 0.533     & 0.580     &85    \\
& CFNet \cite{valmadre2017end}                   & 0.611       & 0.530     & 0.568     &75    \\
& BACF \cite{galoogahi2017learning}              & 0.656       & 0.570     & 0.621     &35    \\
& ECO-hc \cite{danelljan2017eco}                 & 0.652       & 0.592     & 0.643     &60    \\
& MKCFup \cite{tang2018high}                     & 0.641       & -         & -         &150   \\
& MCCT-H \cite{wang2018multi}                    & 0.664       & -         & 0.642     &45    \\ \hline
\multirow{9}{*}{\rotatebox{90}{SiamFC-based Trackers}}
& SiamFC \cite{bertinetto2016fully}              & 0.607       & 0.516     & 0.582     &86   \\
& DSiamM \cite{guo2017learning}                  & 0.656       & -         & -         &25    \\
& RASNet \cite{wang2018learning}                 & 0.670       & -         & 0.642     &83    \\
& SiamRPN \cite{li2018high}                      & 0.658       & 0.592     & 0.637     &200   \\
& SA-Siam \cite{he2018twofold}                   & 0.677       & 0.610     & 0.657     &50    \\
& StructSiam \cite{zhang2018structured}          & 0.637       & -         & 0.621     &45    \\
& MemTrack \cite{yang2018mem}               & 0.642       & -         & 0.626     &50    \\ 
& DaSiamRPN \cite{zhu2018distractor}             & 0.656       & 0.602     & 0.658    &160   \\
& Siam-BM \cite{he2018towards}                    & \textcolor{blue}{0.684}       & \textcolor{blue}{0.617}     & \textcolor{blue}{0.662} & 48 \\ \hline
\multirow{4}{*}{\rotatebox{90}{Misc.}}
& EAST \cite{huang2017learning}                  & 0.638       & -         & 0.629     &159   \\
& PTAV \cite{fan2017parallel}                    & 0.663       & 0.581     & 0.635     &25    \\
& ACT \cite{chen2018real}                        & 0.657       & -         & 0.625     &30    \\
& RT-MDNet \cite{jung18realtime}                 & -           & -         & 0.650     &46    \\ \hline \hline

& Ours                                           & \textbf{\textcolor{red}{0.693}}  & \textbf{\textcolor{red}{0.653}}     & \textbf{\textcolor{red}{0.687}}  & 120   \\ \hline            
\end{tabular}
\end{center}
\caption{Comparison with state-of-the-art real-time trackers on OTB dataset. Trackers are grouped into CF-based methods, SiamFC-based methods and miscellaneous. Numbers in  \textcolor{red}{red} and \textcolor{blue}{blue} are the best and the second best results, respectively.}
\label{table:otb_cmp}
\vspace{-4mm}
\end{table}

\subsection{Analysis of the FM Stage} \label{section:experiment_fm}

\noindent\textbf{Multi-Layer Feature Fusion:} 
The FM stage takes regional features cropped from the shared backbone network as inputs. Generally speaking, deep features are rich in high-level semantic information and shallow features are rich in low-level appearance information. As suggested in many previous works \cite{tao2016sint, wang2015visual, song2017crest, bhat2018unveiling}, multi-layer features can be fused to achieve better performance. We follow this common practice and use \textit{conv2} + \textit{conv4} features for the FM stage. To demonstrate the advantage of multi-layer feature fusion, we compare the performance of SPM-Tracker with alternative implementations which only use single layer features. We train and test models which use \textit{conv2}, \textit{conv3}, or \textit{conv4} only. On OTB-100 benchmark, these three models achieve AUC scores of 0.666, 0.675, and 0.676, respectively, while our final model using \textit{conv2} + \textit{conv4} achieves an AUC of 0.687. This experiment demonstrates that the FM stage benefits from multi-layer feature fusion.

\noindent\textbf{Replace Cross-Correlation Layer with the Relation Network:} 
An important innovation that contributes to the high efficiency of SiamFC tracker is the cross-correlation layer that achieves dense and efficient sliding-window evaluation in the search region. Almost all SiamFC-based trackers have followed this usage. We also use cross-correlation layer in our CM stage for similarity matching and box regression. But in the FM stage, there are much fewer candidate boxes scattered in the search region. There is not much advantage in using cross-correlation operation. Therefore, we replace the cross-correlation layer with a more powerful relation network as described in \cite{yang2018learning}. Experimental results verify our design choice. When models are trained without the GT strategy, using cross-correlation layer in the FM stage results in an AUC of 0.647 on OTB-100, which is slightly better than the single stage baseline SiamRPN (0.643), but is notably inferior to the relation-network-based model (0.670). In addition, when GT is adopted, the AUC score of the cross-correlation-based model is 0.655 while that of the relation-network-based model is 0.687.

\subsection{Comparison with State-of-the-Arts}

    \begin{figure}[]
        \subfigure{%
            \begin{minipage}{0.230\textwidth}
            \centering
            \includegraphics[width=1.0\textwidth]{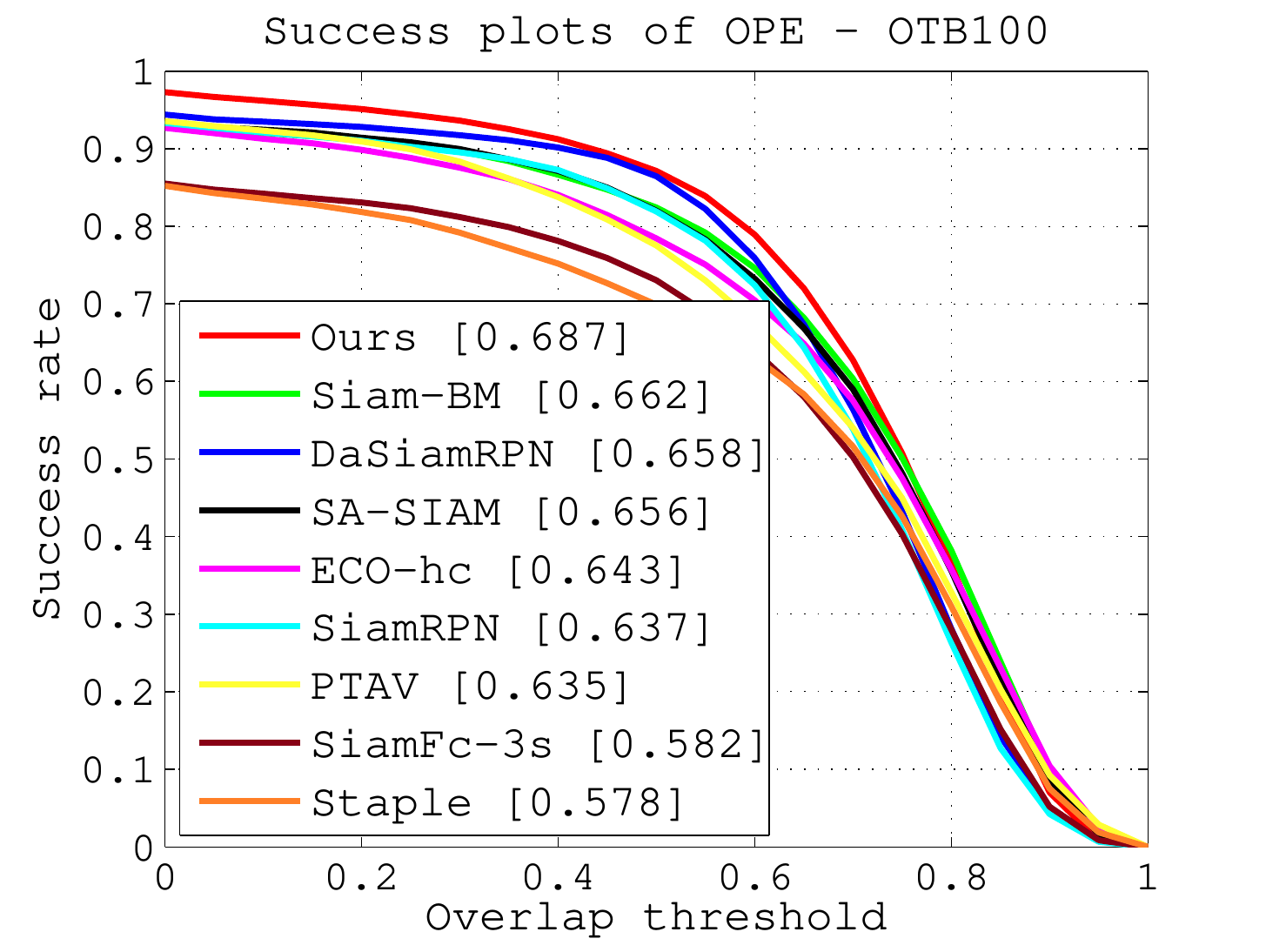}
             \end{minipage}
        }
        \subfigure{%
            \begin{minipage}{0.230\textwidth}
            \centering
            \includegraphics[width=1.0\textwidth]{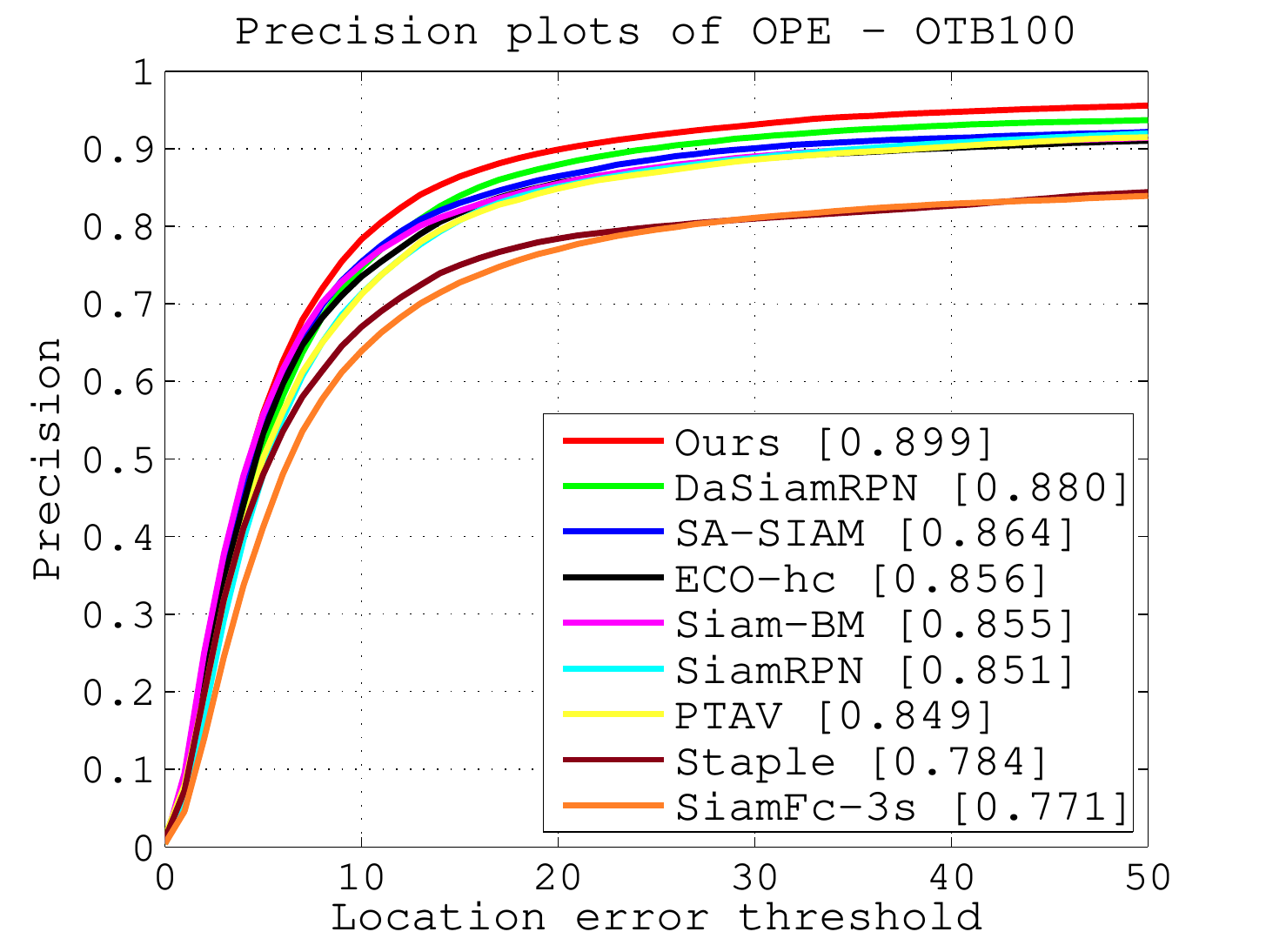}
             \end{minipage}
        }
        \caption{The success plot and precision plot on OTB-100.}
        \label{fig:otb_plot}
        \vspace{-4mm}
    \end{figure}

\noindent\textbf{Evaluation on OTB:} 
Our SPM-Tracker is first compared with the state-of-the-art real-time trackers on OTB 2013/50/100 benchmarks. The detailed AUC scores are summarized in Table \ref{table:otb_cmp}. Due to space limitation, we only show the success plot and the precision plot of one pass evaluation (OPE) on OTB-100 in Fig. \ref{fig:otb_plot}. The SPM-Tracker outperforms other real-time trackers on all three OTB benchmarks by a large margin. 

\begin{figure}[t]
    \begin{center}
        \includegraphics[width=1.0\linewidth]{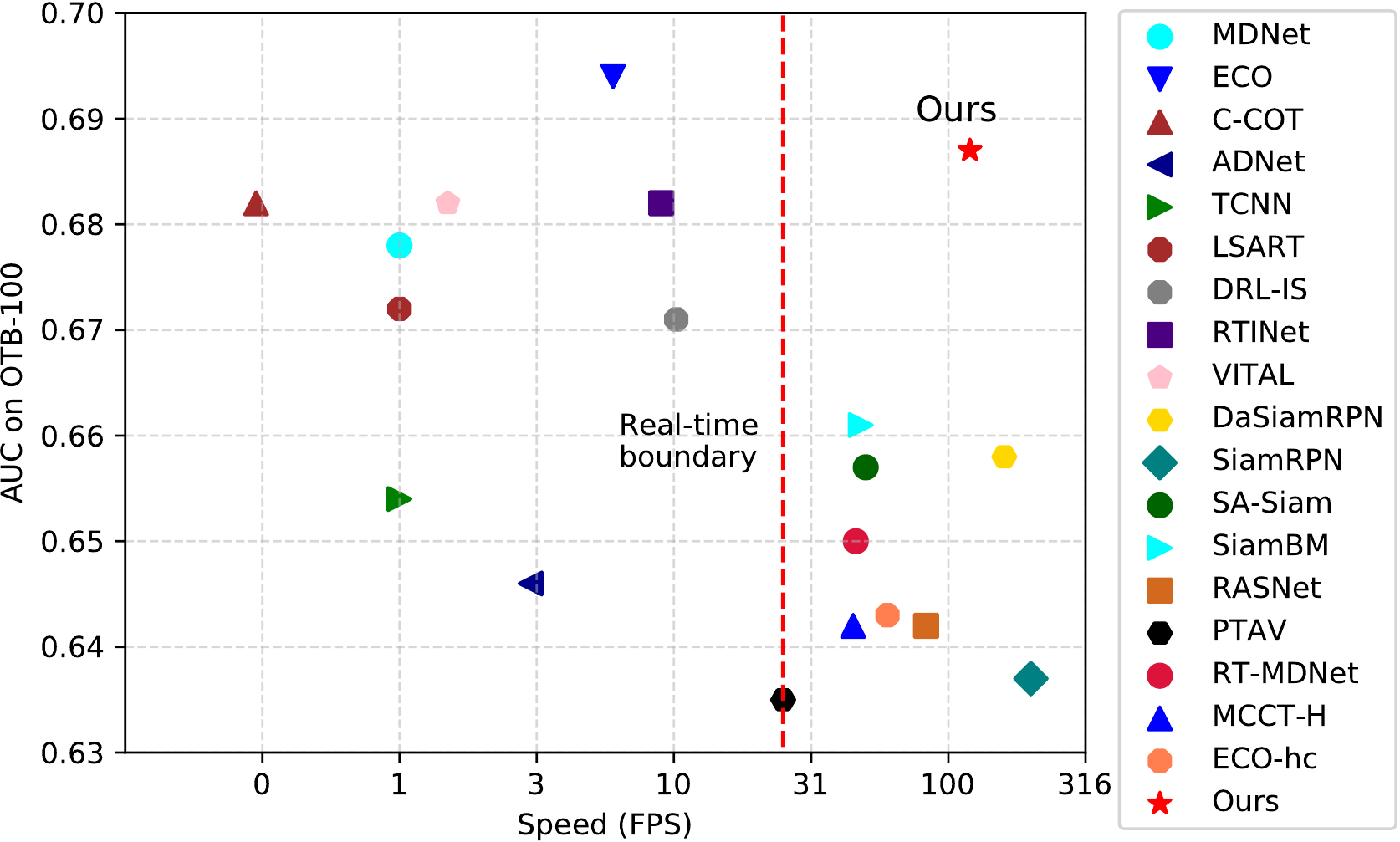}
    \end{center}
    \vspace{-4mm}
    \caption{Performance-speed trade-off of top-performing trackers on OTB-100 benchmark. The speed axis is logarithmic.}
    \label{fig:otb_cmp}
    \vspace{-2mm}
\end{figure}
    
We also compare SPM-Tracker with some non-real-time top-performing trackers, including C-COT \cite{danelljan2016beyond}, ECO \cite{danelljan2017eco}, MDNet \cite{nam2016learning}, ADNet \cite{yoo2017action}, TCCN \cite{nam2016modeling}, LSART \cite{sun2018learning}, VITAL \cite{song2018vital}, RTINet \cite{yao2018joint}, and DRL-IS \cite{ren2018deep}. The AUC score vs. speed curve on OTB-100 is shown in Fig. \ref{fig:otb_cmp}. SPM-Tracker strikes a very good balance between tracking performance and inference speed. 

\begin{table}[]
\small
\setlength{\tabcolsep}{4pt}
\centering
\begin{tabular}{|l|ccc|ccc|l|}
\hline
\multirow{2}{*}{Tracker} & \multicolumn{3}{c|}{VOT-16} & \multicolumn{3}{c|}{VOT-17} & \multirow{2}{*}{FPS} \\ 
                         & A       & R       & EAO     & A       & R       & EAO     &                        \\ \hline
CREST                    & 0.51    & 0.25    & 0.283   & -       & -       & -       & 1                      \\ 
MDNet                    & 0.54    & 0.34    & 0.257   & -       & -       & -       & 1                      \\
C-COT                    & 0.54    & 0.24    & 0.331   & -       & -       & -       & 0.3                    \\
LSART                    & -       & -       & -       & 0.49    & \textcolor{blue}{0.22}    & 0.323   & 1                      \\
ECO                      & 0.55    & \textbf{\textcolor{red}{0.20}}    & 0.375   & 0.48    & 0.27    & 0.280   & 8                      \\ 
UPDT                     & -       & -       & -       & 0.53    & \textbf{\textcolor{red}{0.18}}    & \textbf{\textcolor{red}{0.378}}   & -                      \\ \hline
SiamFC                   & 0.53    & 0.46    & 0.235   & 0.50    & 0.59    & 0.188   & 86                     \\ 
Staple                   & 0.54    & 0.38    & 0.295   & 0.52    & 0.69    & 0.169   & 80                     \\
ECO-hc                   & 0.54    & 0.30    & 0.322   & 0.49    & 0.44    & 0.238   & 60                     \\
SA-Siam                  & 0.54    & -       & 0.291   & 0.50    & 0.46    & 0.236   & 50                     \\
Siam-BM                  & -       & -       & -       & 0.56    & 0.26    & 0.335   & 32                     \\
SiamRPN                  & 0.56    & 0.26    & 0.344   & 0.49    & 0.46    & 0.244   & 200                    \\
DaSiamRPN                & \textcolor{blue}{0.61}    & 0.22    & 0.411   & \textcolor{blue}{0.56}    & 0.34    & 0.326   & 160                    \\ \hline \hline
Ours                    & \textbf{\textcolor{red}{0.62}}    & \textcolor{blue}{0.21}    & \textbf{\textcolor{red}{0.434}}   & \textbf{\textcolor{red}{0.58}}    & 0.30    & \textcolor{blue}{0.338}   & 120                    \\ \hline 
\end{tabular}
\hspace{8pt}
\caption{Comparison with state-of-the-art trackers on VOT benchmark. Both non-real-time methods (top rows) and real-time methods (bottom rows) are included. ``A'' and ``R'' denote accuracy and robustness. EAO stands for expected average overlap. The numbers in \textcolor{red}{red} and \textcolor{blue}{blue} indicate the best and the second best results.}
\label{table:cmp_vot}
\vspace{-2mm}
\end{table}

\noindent\textbf{Evaluation on VOT:} 
SPM-Tracker is evaluated on two VOT benchmark datasets, VOT-16 and VOT-17. Table \ref{table:cmp_vot} shows the comparison with almost all the top-performing trackers despite their speed. Among the real-time trackers, SPM-Tracker is by far the best performing one with superior accuracy and EAO. Even when compared with non-real-time trackers, SPM-Tracker achieves the best accuracy and the EAO performance is among the best. 

\begin{table}[]
\small
\setlength{\tabcolsep}{4pt}
\centering
\begin{tabular}{|l|ccc|l|}
\hline
\multirow{2}{*}{Tracker} & \multicolumn{3}{c|}{AUC of success plot (OPE)}  & \multirow{2}{*}{FPS} \\ 
            & TrackingNet & LaSOT-all & LaSOT-test &  \\ \hline 
MDNet       & 0.606       & 0.413     & 0.397      & 1   \\ 
VITAL       & -           & 0.412     & 0.390      & 1   \\ 
CSRDCF      & 0.534       & 0.263     & 0.244      & 13  \\ 
ECO         & 0.554       & 0.340     & 0.324      & 8   \\ 
ECO-hc      & 0.541       & 0.311     & 0.304      & 60  \\ 
SiamFC      & 0.571       & 0.358     & 0.336      & 86  \\ 
CFNet       & 0.578       & 0.296     & 0.275      & 75  \\ \hline \hline
SiamFC      & 0.602       & 0.378     & 0.364      & 120 \\ 
SiamRPN     & \textcolor{blue}{0.647}           & \textcolor{blue}{0.457}     & \textcolor{blue}{0.432}      & 160 \\ 
SPM-Tracker & \textbf{\textcolor{red}{0.712}}       & \textbf{\textcolor{red}{0.485}}     & \textbf{\textcolor{red}{0.471}}      & 120 \\ \hline
\end{tabular}
\vspace{4pt}
\caption{Comparison with state-of-the-art trackers on TrackingNet and LaSOT. The last three trackers are implemented by ourselves. The numbers in \textcolor{red}{red} and \textcolor{blue}{blue} indicate the best and the second best results.}
\label{table:cmp_large_scale}
\end{table}

\noindent \textbf{Evaluation on large-scale datasets:}
TrackingNet \cite{muller2018trackingnet} and LaSOT \cite{fan2018lasot} are two large-scale datasets for visual tracking. We evaluate our trackers on these two datasets. The results are shown in Table \ref{table:cmp_large_scale}. Our SPM-Tracker performs favorably against state-of-the-art methods in three benchmarks. For fair comparisons, we also implement two Siamese network based methods: SiamFC \cite{bertinetto2016fully} and SiamRPN \cite{li2018high}. Both of them are trained on the same datasets as our SPM-Tracker. The last three rows in Table \ref{table:cmp_large_scale} show that our method achieves a noticeable improvement over SiamFC and SiamRPN. 

\noindent \textbf{Excluding extra data:} Compared with DaSiamRPN \cite{zhu2018distractor}, our tracker has used two more datasets (Cityperson \cite{zhang2017citypersons} and WiderFace \cite{yang2016wider}) in training. For fair comparison, we have also trained a model excluding these two datasets. The AUC on OTB-100 slightly drops to 0.671, but still outperforms DaSiamRPN and Siam-BM. The EAO on VOT-16 becomes 0.432 and that on VOT-17 slightly increases to 0.347.


\subsection{Qualitative Results}

    \begin{figure}
    \begin{center}
    \includegraphics[width=1.0\linewidth]{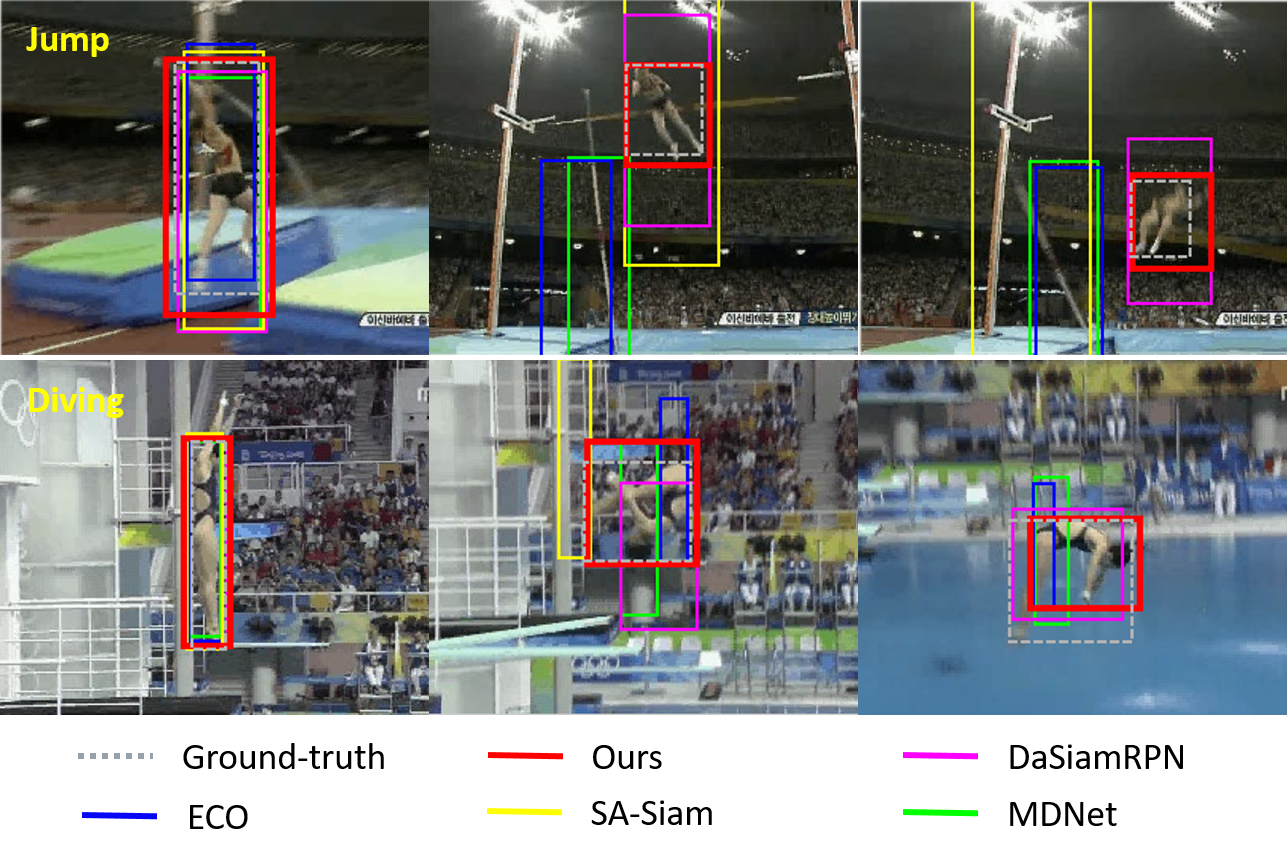}
    \vspace{-4mm}
    \end{center}
       \caption{Visualization of three successful tracking sequences from OTB-100.}
    \label{fig:visualization}
    \vspace{-2mm}
    \end{figure}

\noindent\textbf{Successful Cases: }
In Fig. \ref{fig:visualization}, we visualize three successful tracking cases, including the very challenging \textit{jump} and \textit{diving} sequences. Owing to the robustness of the CM stage, our tracker is able to detect targets with huge deformation. The region proposal branch allows SPM-Tracker to fit to the varying object shapes. In these two sequences, some of the best trackers such as ECO \cite{danelljan2017eco} and MDNet \cite{nam2016learning} also fail. DaSiamRPN \cite{zhu2018distractor} barely follows the target, but the box locations are less precise. This demonstrates the advantage of our two-stage box refinement.

\begin{figure}
\begin{center}
\includegraphics[width=1.0\linewidth]{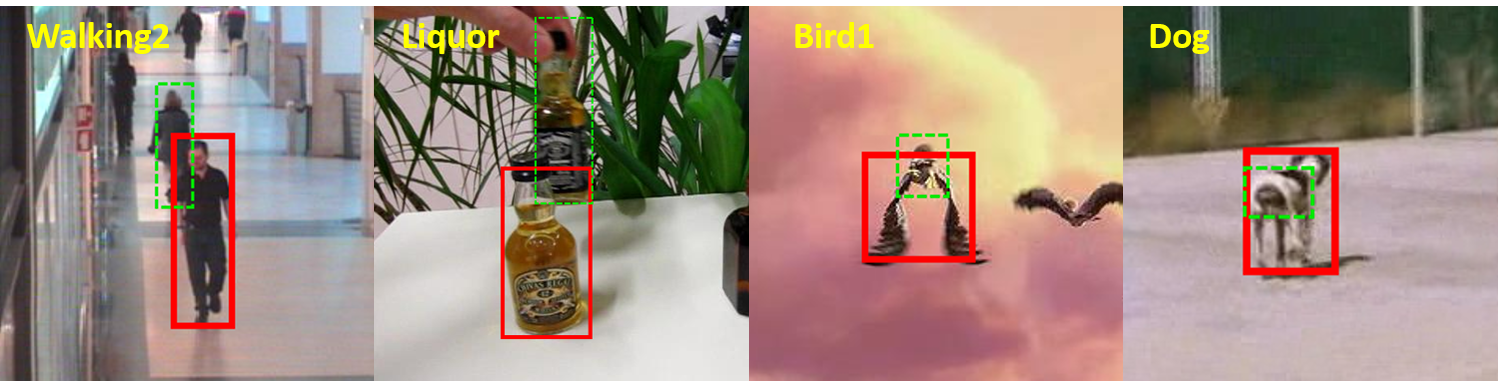}
\caption{Visualization of failure cases. The \textcolor{green}{green} box is ground-truth and the \textcolor{red}{red} box is our tracking result.}
\label{fig:failure_cases}
\vspace{-8mm}
\end{center}
\end{figure}

\noindent\textbf{Failure Cases:} 
We observe two types of failures in SPM-Tracker, as shown in Fig. \ref{fig:failure_cases}. In \textit{walking2} and \textit{liquor} sequences, when the target is occluded by a similar object, the tracking box may drift. The other type of failure occurs when the ground-truth target is only a part of an object, as in sequences \textit{bird1} and \textit{dog}. SPM-Tracker seems to have a strong sense of objectness and tends to track the entire object even when the template only contains a part of it.

\section{Conclusion} \label{section:conclusion}

We have presented the design and implementation of a static discriminative tracker named SPM-Tracker. SPM-Tracker adopts a novel series-parallel structure for two-stage matching. Evaluations on OTB and VOT benchmarks show its superior tracking performance. In the future, we plan to explore solutions to the drifting problem when the target is occluded by similar objects. Possible choices include template update and forward-backward verification. We believe that the series-parallel matching framework has great potential and is worthy of further investigation. 

\section*{Acknowledgement}

We acknowledge funding from National Key R\&D Program of China under Grant 2017YFA0700800.

{\small
\bibliographystyle{ieee}
\bibliography{spm_arxiv}

\begin{thebibliography}{10}\itemsep=-1pt

\bibitem{bertinetto2016staple}
L.~Bertinetto, J.~Valmadre, S.~Golodetz, O.~Miksik, and P.~H. Torr.
\newblock Staple: Complementary learners for real-time tracking.
\newblock In {\em CVPR}, pages 1401--1409, 2016.

\bibitem{bertinetto2016fully}
L.~Bertinetto, J.~Valmadre, J.~F. Henriques, A.~Vedaldi, and P.~H. Torr.
\newblock Fully-convolutional siamese networks for object tracking.
\newblock In {\em ECCV}, pages 850--865, 2016.

\bibitem{bhat2018unveiling}
G.~Bhat, J.~Johnander, M.~Danelljan, F.~S. Khan, and M.~Felsberg.
\newblock Unveiling the power of deep tracking.
\newblock In {\em ECCV}, 2018.

\bibitem{bolme2010visual}
D.~S. Bolme, J.~R. Beveridge, B.~A. Draper, and Y.~M. Lui.
\newblock Visual object tracking using adaptive correlation filters.
\newblock In {\em CVPR}, pages 2544--2550, 2010.

\bibitem{chen2018real}
B.~Chen, D.~Wang, P.~Li, S.~Wang, and H.~Lu.
\newblock Real-time ‘actor-critic’tracking.
\newblock In {\em ECCV}, pages 328--345, 2018.

\bibitem{chen2014description}
D.~Chen, Z.~Yuan, G.~Hua, Y.~Wu, and N.~Zheng.
\newblock Description-discrimination collaborative tracking.
\newblock In {\em ECCV}, pages 345--360, 2014.

\bibitem{danelljan2017eco}
M.~Danelljan, G.~Bhat, F.~S. Khan, M.~Felsberg, et~al.
\newblock Eco: Efficient convolution operators for tracking.
\newblock In {\em CVPR}, pages 6931--6939, 2017.

\bibitem{danelljan2015learning}
M.~Danelljan, G.~Hager, F.~Shahbaz~Khan, and M.~Felsberg.
\newblock Learning spatially regularized correlation filters for visual
  tracking.
\newblock In {\em ICCV}, pages 4310--4318, 2015.

\bibitem{danelljan2016beyond}
M.~Danelljan, A.~Robinson, F.~S. Khan, and M.~Felsberg.
\newblock Beyond correlation filters: Learning continuous convolution operators
  for visual tracking.
\newblock In {\em ECCV}, pages 472--488, 2016.

\bibitem{danelljan2014adaptive}
M.~Danelljan, F.~Shahbaz~Khan, M.~Felsberg, and J.~Van~de Weijer.
\newblock Adaptive color attributes for real-time visual tracking.
\newblock In {\em CVPR}, pages 1090--1097, 2014.

\bibitem{fan2018lasot}
H.~Fan, L.~Lin, F.~Yang, P.~Chu, G.~Deng, S.~Yu, H.~Bai, Y.~Xu, C.~Liao, and
  H.~Ling.
\newblock Lasot: A high-quality benchmark for large-scale single object
  tracking.
\newblock {\em arXiv preprint arXiv:1809.07845}, 2018.

\bibitem{fan2017parallel}
H.~Fan and H.~Ling.
\newblock Parallel tracking and verifying: A framework for real-time and high
  accuracy visual tracking.
\newblock In {\em ICCV}, pages 5487--5495, 2017.

\bibitem{galoogahi2017learning}
H.~K. Galoogahi, A.~Fagg, and S.~Lucey.
\newblock Learning background-aware correlation filters for visual tracking.
\newblock In {\em CVPR}, pages 1144--1152, 2017.

\bibitem{gao2014transfer}
J.~Gao, H.~Ling, W.~Hu, and J.~Xing.
\newblock Transfer learning based visual tracking with gaussian processes
  regression.
\newblock In {\em ECCV}, pages 188--203, 2014.

\bibitem{girshick15fastrcnn}
R.~Girshick.
\newblock Fast {R-CNN}.
\newblock In {\em ICCV}, pages 1440--1448, 2015.

\bibitem{guo2017learning}
Q.~Guo, W.~Feng, C.~Zhou, R.~Huang, L.~Wan, and S.~Wang.
\newblock Learning dynamic siamese network for visual object tracking.
\newblock In {\em ICCV}, pages 1763--1771, 2017.

\bibitem{he2018towards}
A.~He, C.~Luo, X.~Tian, and W.~Zeng.
\newblock Towards a better match in siamese network based visual object
  tracker.
\newblock In {\em ECCV Workshop}, 2018.

\bibitem{he2018twofold}
A.~He, C.~Luo, X.~Tian, and W.~Zeng.
\newblock A twofold siamese network for real-time object tracking.
\newblock In {\em CVPR}, pages 4834--4843, 2018.

\bibitem{he2017mask}
K.~He, G.~Gkioxari, P.~Doll{\'a}r, and R.~Girshick.
\newblock Mask r-cnn.
\newblock In {\em ICCV}, pages 2980--2988, 2017.

\bibitem{henriques2012exploiting}
J.~F. Henriques, R.~Caseiro, P.~Martins, and J.~Batista.
\newblock Exploiting the circulant structure of tracking-by-detection with
  kernels.
\newblock In {\em ECCV}, pages 702--715, 2012.

\bibitem{henriques2015high}
J.~F. Henriques, R.~Caseiro, P.~Martins, and J.~Batista.
\newblock High-speed tracking with kernelized correlation filters.
\newblock {\em T-PAMI}, 37(3):583--596, 2015.

\bibitem{huang2017learning}
C.~Huang, S.~Lucey, and D.~Ramanan.
\newblock Learning policies for adaptive tracking with deep feature cascades.
\newblock In {\em ICCV}, pages 105--114, 2017.

\bibitem{huang2017speed}
J.~Huang, V.~Rathod, C.~Sun, M.~Zhu, A.~Korattikara, A.~Fathi, I.~Fischer,
  Z.~Wojna, Y.~Song, S.~Guadarrama, et~al.
\newblock Speed/accuracy trade-offs for modern convolutional object detectors.
\newblock In {\em CVPR}, pages 7310--7311, 2017.

\bibitem{jung18realtime}
I.~Jung, J.~Son, M.~Baek, and B.~Han.
\newblock Real-time mdnet.
\newblock In {\em ECCV}, pages 83--98, 2018.

\bibitem{kalal2011TLD}
Z.~Kalal, K.~Mikolajczyk, J.~Matas, et~al.
\newblock Tracking-learning-detection.
\newblock {\em T-PAMI}, 34(7):1409, 2012.

\bibitem{krizhevsky2012imagenet}
A.~Krizhevsky, I.~Sutskever, and G.~E. Hinton.
\newblock Imagenet classification with deep convolutional neural networks.
\newblock In {\em NIPS}, pages 1097--1105, 2012.

\bibitem{li2018high}
B.~Li, J.~Yan, W.~Wu, Z.~Zhu, and X.~Hu.
\newblock High performance visual tracking with siamese region proposal
  network.
\newblock In {\em CVPR}, pages 8971--8980, 2018.

\bibitem{li2014scale}
Y.~Li and J.~Zhu.
\newblock A scale adaptive kernel correlation filter tracker with feature
  integration.
\newblock In {\em ECCV}, pages 254--265, 2014.

\bibitem{lin2017focal}
T.-Y. Lin, P.~Goyal, R.~Girshick, K.~He, and P.~Doll\'{a}r.
\newblock {Focal Loss for Dense Object Detection}.
\newblock In {\em ICCV}, pages 2980--2988, 2017.

\bibitem{lin2014microsoft}
T.-Y. Lin, M.~Maire, S.~Belongie, J.~Hays, P.~Perona, D.~Ramanan,
  P.~Doll{\'a}r, and C.~L. Zitnick.
\newblock Microsoft coco: Common objects in context.
\newblock In {\em ECCV}, pages 740--755, 2014.

\bibitem{lu2018deep}
X.~Lu, C.~Ma, B.~Ni, X.~Yang, I.~Reid, and M.-H. Yang.
\newblock Deep regression tracking with shrinkage loss.
\newblock In {\em ECCV}, pages 353--369, 2018.

\bibitem{ma2015hierarchical}
C.~Ma, J.-B. Huang, X.~Yang, and M.-H. Yang.
\newblock Hierarchical convolutional features for visual tracking.
\newblock In {\em ICCV}, pages 3074--3082, 2015.

\bibitem{ma2015long}
C.~Ma, X.~Yang, C.~Zhang, and M.-H. Yang.
\newblock Long-term correlation tracking.
\newblock In {\em CVPR}, pages 5388--5396, 2015.

\bibitem{muller2018trackingnet}
M.~Muller, A.~Bibi, S.~Giancola, S.~Alsubaihi, and B.~Ghanem.
\newblock Trackingnet: A large-scale dataset and benchmark for object tracking
  in the wild.
\newblock In {\em ECCV}, 2018.

\bibitem{nam2016modeling}
H.~Nam, M.~Baek, and B.~Han.
\newblock Modeling and propagating cnns in a tree structure for visual
  tracking.
\newblock {\em arXiv preprint}, 2016.

\bibitem{nam2016learning}
H.~Nam and B.~Han.
\newblock Learning multi-domain convolutional neural networks for visual
  tracking.
\newblock In {\em CVPR}, pages 4293--4302, 2016.

\bibitem{real2017youtube}
E.~Real, J.~Shlens, S.~Mazzocchi, X.~Pan, and V.~Vanhoucke.
\newblock Youtube-boundingboxes: A large high-precision human-annotated data
  set for object detection in video.
\newblock In {\em CVPR}, pages 7464--7473, 2017.

\bibitem{ren2018deep}
L.~Ren, X.~Yuan, J.~Lu, M.~Yang, and J.~Zhou.
\newblock Deep reinforcement learning with iterative shift for visual tracking.
\newblock In {\em ECCV}, pages 684--700, 2018.

\bibitem{ren2015faster}
S.~Ren, K.~He, R.~Girshick, and J.~Sun.
\newblock Faster r-cnn: Towards real-time object detection with region proposal
  networks.
\newblock In {\em NIPS}, pages 91--99, 2015.

\bibitem{russakovsky2015imagenet}
O.~Russakovsky, J.~Deng, H.~Su, J.~Krause, S.~Satheesh, S.~Ma, Z.~Huang,
  A.~Karpathy, A.~Khosla, M.~Bernstein, et~al.
\newblock Imagenet large scale visual recognition challenge.
\newblock {\em IJCV}, 115(3):211--252, 2015.

\bibitem{song2017crest}
Y.~Song, C.~Ma, L.~Gong, J.~Zhang, R.~W. Lau, and M.-H. Yang.
\newblock Crest: Convolutional residual learning for visual tracking.
\newblock In {\em ICCV}, pages 2574--2583, 2017.

\bibitem{song2018vital}
Y.~Song, C.~Ma, X.~Wu, L.~Gong, L.~Bao, W.~Zuo, C.~Shen, R.~Lau, and M.-H.
  Yang.
\newblock Vital: Visual tracking via adversarial learning.
\newblock In {\em CVPR}, 2018.

\bibitem{sun2018learning}
C.~Sun, H.~Lu, and M.-H. Yang.
\newblock Learning spatial-aware regressions for visual tracking.
\newblock In {\em CVPR}, pages 8962--8970, 2018.

\bibitem{tang2018high}
M.~Tang, B.~Yu, F.~Zhang, and J.~Wang.
\newblock High-speed tracking with multi-kernel correlation filters.
\newblock In {\em CVPR}, pages 4874--4883, 2018.

\bibitem{tao2016sint}
R.~Tao, E.~Gavves, and A.~W.~M. Smeulders.
\newblock Siamese instance search for tracking.
\newblock In {\em CVPR}, pages 1420--1429, 2016.

\bibitem{valmadre2017end}
J.~Valmadre, L.~Bertinetto, J.~Henriques, A.~Vedaldi, and P.~H. Torr.
\newblock End-to-end representation learning for correlation filter based
  tracking.
\newblock In {\em CVPR}, pages 5000--5008, 2017.

\bibitem{wang2015visual}
L.~Wang, W.~Ouyang, X.~Wang, and H.~Lu.
\newblock Visual tracking with fully convolutional networks.
\newblock In {\em ICCV}, pages 3119--3127, 2015.

\bibitem{wang2017large}
M.~Wang, Y.~Liu, and Z.~Huang.
\newblock Large margin object tracking with circulant feature maps.
\newblock In {\em CVPR}, pages 21--26, 2017.

\bibitem{wang2015transferring}
N.~Wang, S.~Li, A.~Gupta, and D.-Y. Yeung.
\newblock Transferring rich feature hierarchies for robust visual tracking.
\newblock {\em arXiv preprint}, 2015.

\bibitem{wang2018multi}
N.~Wang, W.~Zhou, Q.~Tian, R.~Hong, M.~Wang, and H.~Li.
\newblock Multi-cue correlation filters for robust visual tracking.
\newblock In {\em CVPR}, pages 4844--4853, 2018.

\bibitem{wang2018learning}
Q.~Wang, Z.~Teng, J.~Xing, J.~Gao, W.~Hu, and S.~Maybank.
\newblock Learning attentions: residual attentional siamese network for high
  performance online visual tracking.
\newblock In {\em CVPR}, pages 4854--4863, 2018.

\bibitem{yang2018learning}
F.~S.~Y. Yang, L.~Zhang, T.~Xiang, P.~H. Torr, and T.~M. Hospedales.
\newblock Learning to compare: Relation network for few-shot learning.
\newblock In {\em CVPR}, 2018.

\bibitem{yang2016wider}
S.~Yang, P.~Luo, C.-C. Loy, and X.~Tang.
\newblock Wider face: A face detection benchmark.
\newblock In {\em CVPR}, pages 5525--5533, 2016.

\bibitem{yang2018mem}
T.~Yang and A.~B. Chan.
\newblock Learning dynamic memory networks for object tracking.
\newblock In {\em ECCV}, 2018.

\bibitem{yao2018joint}
Y.~Yao, X.~Wu, L.~Zhang, S.~Shan, and W.~Zuo.
\newblock Joint representation and truncated inference learning for correlation
  filter based tracking.
\newblock In {\em ECCV}, pages 552--567, 2018.

\bibitem{yoo2017action}
S.~Yoo, K.~Yun, J.~Y. Choi, K.~Yun, and J.~Choi.
\newblock Action-decision networks for visual tracking with deep reinforcement
  learning.
\newblock In {\em CVPR}, pages 1349--1358, 2017.

\bibitem{zhang2018LRVNT}
L.~W. J. Q. H.~L. Yunhua~Zhang, Dong~Wang.
\newblock Learning regression and verification networks for long-term visual
  tracking.
\newblock In {\em arXiv preprint arXiv:1809.04320}, 2018.

\bibitem{zhang2017citypersons}
S.~Zhang, R.~Benenson, and B.~Schiele.
\newblock Citypersons: A diverse dataset for pedestrian detection.
\newblock In {\em CVPR}, pages 3213--3221, 2017.

\bibitem{zhang2018structured}
Y.~Zhang, L.~Wang, J.~Qi, D.~Wang, M.~Feng, and H.~Lu.
\newblock Structured siamese network for real-time visual tracking.
\newblock In {\em ECCV}, pages 351--366, 2018.

\bibitem{zhu2018distractor}
Z.~Zhu, Q.~Wang, B.~Li, W.~Wu, J.~Yan, and W.~Hu.
\newblock Distractor-aware siamese networks for visual object tracking.
\newblock In {\em ECCV}, pages 103--119, 2018.

\end{thebibliography}
}

\end{document}